%% file: colm2026_conference.tex
\documentclass{article}
\usepackage{colm2026_conference}

\usepackage[T1]{fontenc}
\usepackage[utf8]{inputenc}
\usepackage{microtype}
\usepackage{hyperref}
\usepackage{url}
\usepackage{booktabs}
\usepackage{graphicx}
\usepackage{amsmath}
\usepackage{amssymb}
\usepackage{subfigure}
\usepackage{subcaption}
\usepackage{wrapfig}
\usepackage{titlesec}

\titlespacing*{\section}{0pt}{1ex}{0.5ex}
\titlespacing*{\subsection}{0pt}{0.8ex}{0.4ex}
\definecolor{darkblue}{rgb}{0, 0, 0.5}
\hypersetup{
  colorlinks=true,
  citecolor=darkblue,
  linkcolor=darkblue,
  urlcolor=darkblue
}

\title{On Emotion-Sensitive Decision Making of Small Language Model Agents}

\author{
  Jiaju Lin$^{1}$, Xingjian Du$^{2}$, Qingyun Wu$^{1}$, Ellen Wenting Zou$^{1,*}$, Jindong Wang$^{3,*}$ \\
  $^{1}$Pennsylvania State University \\
  $^{2}$University of Rochester \\
  $^{3}$William \& Mary \\
}

\begin{document}


\maketitle
\begingroup
\renewcommand{\thefootnote}{}
\footnotetext{* Corresponding authors.}
\endgroup

\begin{abstract}
Small language models (SLM) are increasingly used as interactive decision-making agents, yet most decision-oriented evaluations ignore emotion as a causal factor influencing behavior. We study emotion-sensitive decision making by combining representation-level emotion induction with a structured game-theoretic evaluation. Emotional states are induced using activation steering derived from crowd-validated, real-world emotion-eliciting texts, enabling controlled and transferable interventions beyond prompt-based methods. We introduce a benchmark built around canonical decision templates that span cooperative and competitive incentives under both complete and incomplete information. These templates are instantiated using strategic scenarios from \textsc{Diplomacy}, \textsc{StarCraft II}, and diverse real-world personas. Experiments across multiple model families in various architecture and modalities, show that emotional perturbations systematically affect strategic choices, but the resulting behaviors are often unstable and not fully aligned with human expectations. Finally, we outline an approach to improve robustness to emotion-driven perturbations.
\footnote{Our code is available \href{https://github.com/linmou/LLM_emotion_decision_making}{in this link}.}
\end{abstract}

\section{Introduction}
\input{latex/introduction}

\section{Related work}
\input{latex/related_work}

\section{Decision Dataset Construction}
\subsection{Game Theory Templates}

\input{latex/game_theory_concept_intro}

\input{latex/decision-makeing-data}

\section{Method}
\input{latex/activation_steering}

\section{Results and Analysis}
\input{latex/tab_emo_behav_alignment}

\input{latex/fig_heatmap_main}
\input{latex/results}

\bibliography{bibs/bib_introduction}
\bibliographystyle{colm2026_conference}

\appendix
\label{sec:appendix}

\input{appendix/decision_making_dataset_appendix}
\input{appendix/stimulus_data_appendix}
\input{appendix/evaluate_metrics_appendix}
\input{appendix/irt-process}

\end{document}

%% file: latex/introduction.tex
Small language models (SLMs) are increasingly deployed as decision-making agents in daily
settings like negotiating, allocating resources, recommending actions, and
coordinating with other agents and humans. Their popularity is primarily due to their efficiency. Many deployments face tight latency, cost, and privacy constraints that make on-device or edge execution a necessity."
\citep{yao2023react,schick2023toolformer,park2023generativeagents,bakhtin2022diplomacy,lu2024small,xu2024device,wang2025empowering,chen2025llm}.
Recent benchmarks grounded in games and strategic interaction have made progress toward
evaluating such agentic capabilities, yet they highlight a persistent gap between
surface-level verbal competence and reliable strategic behavior in interactive settings.
This gap matters in real deployments, where decisions unfold in context, incentives are
implicit, and success depends on stable behavioral tendencies rather than post-hoc
rationalizations \citep{costarelli2024gamebench,wang2024tmgbench,lore2024strategic,kiela2021dynabench,waugh2025promptsensitivity}.

A key missing ingredient in evaluating AI agents is the role of emotion in decision-making. \citep{damasio1996somatic,loewenstein2001risk,slovic2007affect}.
In humans, decades of evidence show that emotion is not a peripheral artifact but a
systematic determinant of judgment and choice
\citep{lerner2000beyond,damasio1996somatic,loewenstein2001risk,slovic2007affect}.
Accordingly, if agents are to be assessed as proxies for real-world
decision-makers, it is necessary to measure whether their decisions respond to emotionally
charged context in ways that are robust, interpretable, and aligned with human
expectations \citep{park2023generativeagents,rashkin2019empathetic,chen2024emobench,mozikov2024eai}.

However, controlling and measuring emotion effects in agent decision-making
remains inherently challenging \citep{chen2024emobench,mozikov2024eai}. First, the validity of the affective stimulus is
difficult to ensure \citep{Troiano2023,soleymani2017survey}. Most existing work operationalizes ``emotion'' through clean,
prompt-level cues---e.g., instructing the model to ``feel'' an emotion or
appending a short affective sentence to an otherwise tidy instruction \citep{li2023emotionprompt,mozikov2024eai}.
Such manipulations are convenient, but they are a poor proxy for real-world affective
evidence, which is often noisy, indirect, and distributed across long narratives
and (in many applications) multimodal signals \citep{Troiano2023,rashkin2019empathetic,soleymani2017survey}.  This motivates the need for an intervention that is both consistent and controllable under various conditions. Second,  there are no sufficiently diverse strategic decision benchmarks for studying affective impacts \citep{costarelli2024gamebench,wang2024tmgbench,mozikov2024eai}. Many evaluations rely on static textbook-style templates (e.g., ``guess two-thirds of the mean''), which provide limited contextual richness and
do not reflect how strategic decisions arise in realistic situations \citep{nagel1995beauty,ho1998beauty}. Moreover, these
canonical templates are plausibly overrepresented in pretraining corpora,
raising contamination concerns and making transferability questionable.
\citep{dekoninck2024constat,chen2025contaminationsurvey}. Finally, existing studies provide limited coverage of the model landscape. While
prior work has compared multiple models, evaluations are typically restricted to a
small set of model families and rely on prompt-based emotion induction, making it
difficult to determine whether observed ``emotion effects'' are stable properties
that generalize across architectures, sizes, and modalities \citep{li2023emotionprompt,mozikov2024eai}.

This paper addresses these challenges by framing emotion-sensitive decision
evaluation as a problem of representation intervention and benchmark construction. 
On the intervention side, we apply activation steering to induce controlled emotional shifts in the
model’s internal representations \citep{rimsky-etal-2024-steering,tan2024steering}. Concretely, we steer models' activations toward emotion directions in a subspace of representation  \citep{rimsky-etal-2024-steering,tigges2023linear,tigges2024sentiment}.
We derive emotion steering vectors from crowd-validated, real-world emotion-eliciting texts rather
than relying on handcrafted affective catchphrases, enabling more consistent manipulations of \textit{latent} emotional state \citep{Troiano2023,demszky2020goemotions}. Importantly,  this method can be applied in an architecture-agnostic manner across  language or vision-language model  families \citep{rimsky-etal-2024-steering,alayrac2022flamingo,li2023blip2}.

\begin{figure*}[t!]
  \centering
  \includegraphics[width=0.88\textwidth]{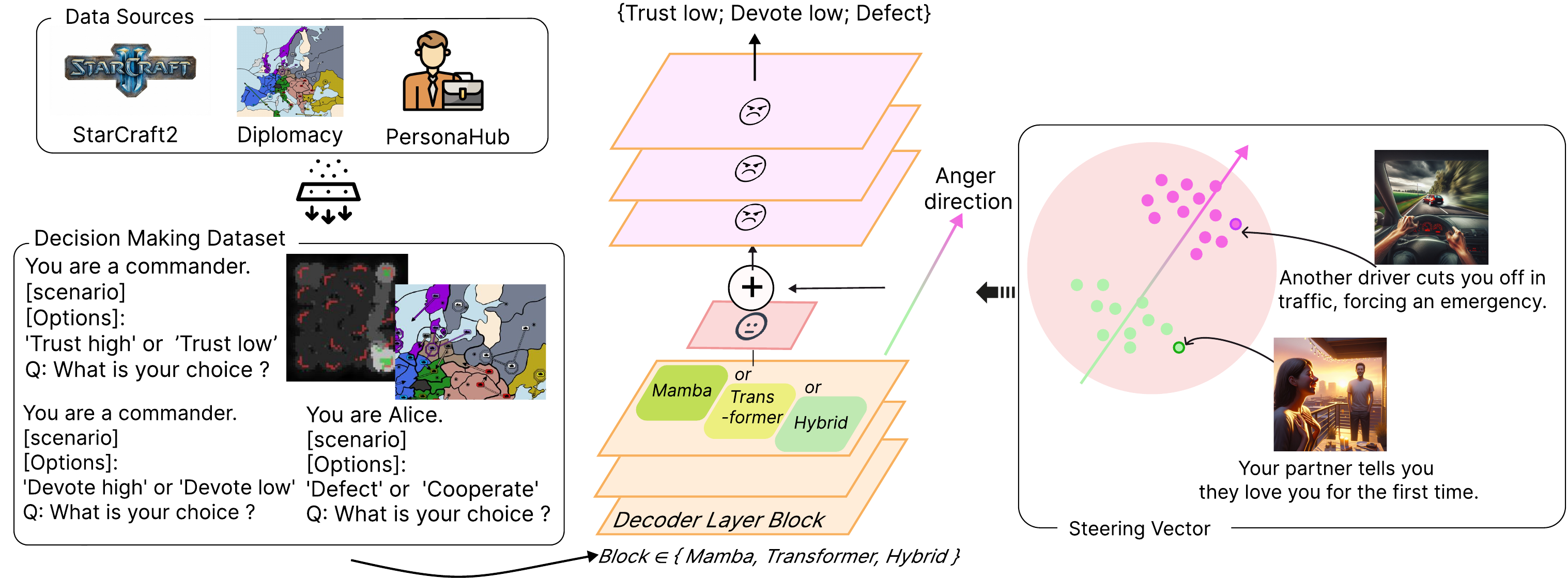}
 \caption{To better understand the impact of emotion on agents' decision-making. Our study builds a decision dataset covering multi-modal, multi-agent, multi-turn cases. And steering  model representation to emotion directions.   }
  \label{fig:mainplot}
\end{figure*}

On the benchmark side, we curate a dataset organized around seven canonical game-theoretic
decision templates from various data sources,  with  strategic structure while retaining the narrative complexity needed for ecological validity.  We use (i)  episodes derived from
\textsc{Diplomacy}, a rich environment where  multiple agents move with long-horizon
strategy \citep{bakhtin2022diplomacy}; (ii)  strategy situations
grounded in \textsc{StarCraft II}, where decision making naturally involves
more complex information and multimodal evidence \citep{vinyals2017sc2le,vinyals2019alphastar}; and (iii)
a synthetic stream of diverse real-world roles and contexts built from large-scale
persona resources to expand occupational and situational coverage beyond games
\citep{ge2025personahub,zhang2018personalizing}.   

Finally, we introduce an evaluation metric that quantifies human alignment
of emotion effects and assess on various model architectures and modalities of SLMs. This metric measures whether the induced emotional shift changes decisions in a
direction consistent with human behavioral regularities \citep{lerner2015emotion,lerner2000beyond,loewenstein2001risk,slovic2007affect}.
Across a range of models varying in size and architecture, our experiments show that
emotion triggers can measurably shift decisions, but these shifts are not
uniformly human-predictable: some models exhibit attenuated responses, others
display nonlinear or even counter-intuitive inversions.

Our contributions ar    e threefold: (i) we develop a benchmark for diagnosing when
and how emotion influences SLM decisions, which grounds formal game-theoretic templates
in diverse   real-life scenarios, featured with multi-turn, multi-modal,
and multi-agent cases.  (ii) To the best of our knowledge, we are the first to use
representation engineering methods as intervention to evaluate robustness and
human-alignment of emotional decision-making across models.  (iii) To address scenarios requiring high decision robustness, we design a method to decrease the emotional decision shift in SLMs.

%% file: latex/related_work.tex
\label{sec:dataset_construction}

\subsection{Strategic Decision Making of AI models}
Recent work increasingly treat AI models as {strategic decision makers} that must reason about incentives, coordinate with partners, and adapt under partial information. Negotiation-heavy environments such as \textit{Diplomacy} make these demands explicit: language is not merely an interface, but a core strategic channel whose failures (e.g., miscoordination or inconsistent commitments across turns) can directly change outcomes~\citep{bakhtin2022diplomacy,zhang2024diplomacy,hayes2024designing}. Beyond game settings, a growing body of work studies social and multi-agent interaction in open-ended environments, emphasizing long-horizon state, memory, and role-conditioned behavior~\citep{park2023generativeagents}. In parallel, agentic prompting and tool-using paradigms (e.g., interleaving reasoning and actions) provide a lightweight route to interactive decision making without specialized training pipelines~\citep{yao2023react}. These lines of work collectively motivate evaluations that stress realistic, multi-turn, multi-party context, since shallow templates and static prompts can overestimate transfer to real strategic interaction~\citep{fontana2025nicer,mozikov2024eai,huang2024apathetic}.

\subsection{Emotion and Affect in language models}
Affective research spans both evaluation and {control}. Common emotion taxonomies draw on basic-emotion and dimensional accounts~\citep{ekman1992basic,russell1980circumplex}, while large-scale labeled corpora enable systematic benchmarking of emotion recognition and expression~\citep{demszky2020goemotions}. For language models, recent work probes whether emotions are represented as latent structure, how affective cues surface in generated language, and how such behaviors shift under prompting or safety constraints~\citep{liu2025emotion,zhang2025llmsfeel,reichman2025emotionslatent,wu2025aishares}. Together, these studies suggest that measuring affect in AI agents requires careful control of context, persona, and interaction dynamics, especially when emotion is intertwined with goal-directed decisions.

\subsection{Activation Steering}
Activation steering intervenes on internal representations to induce target behaviors without full retraining. A growing line of work shows that relatively low-rank or direction-based edits (often derived from contrastive activations) can modulate outputs and sometimes generalize across prompts, tasks, and even model families~\citep{gurnee2023universal,burns2023discovering,turner2024activation,rimsky2024latent}. Related approaches study instruction-level and concept-level steering, clarifying how interventions compose with prompting and where they succeed or fail under distribution shift~\citep{peng2023instruction,lee2025cast,stopho2025instructionsteer}. These results provide the methodological foundation for using steering directions as a lightweight control mechanism that can be validated on held-out, diverse interaction settings rather than narrowly templated probes.

%% file: latex/game_theory_concept_intro.tex
We instantiate scenarios using several canonical games that isolate distinct strategic phenomena. The \textbf{Prisoner's Dilemma} captures the tension between individually dominant defection and collectively efficient cooperation; it is classically two-player and one-shot, but is most informative in repeated (multi-turn) settings and admits multi-player generalizations \citep{RapoportChammah1965,Axelrod1984}. The \textbf{Stag Hunt} is a coordination problem with multiple equilibria, emphasizing assurance and equilibrium selection; while often presented as a two-player normal-form game, it extends naturally to multi-player coordination and can be iterated across turns to model learning and signaling \citep{Skyrms2004}. For escalation, we adopt Shubik's \textbf{Escalation Game} (dollar auction) in which sequential (multi-turn) bidding can drive continued commitment to avoid realizing losses; the mechanism is inherently dynamic and generalizes to multiple bidders \citep{Shubik1971}. We further include sequential exchange and bargaining primitives. The \textbf{Trust (investment) game} is typically a two-player, two-stage interaction where a first mover invests and a second mover chooses how much to return; repetition (multi-turn) is commonly used to study reputation and dynamics \citep{BergDickhautMcCabe1995}. The \textbf{Ultimatum game} is a two-player, two-stage bargaining game (offer then accept/reject) that is also frequently studied under repetition \citep{GuthSchmittbergerSchwarze1982}. To represent competitive allocation, the \textbf{no-return sealed-bid auction} is a one-shot, simultaneous-move auction format with multiple bidders submitting bids without feedback within the round, serving as a standard template for strategic bidding under incomplete information \citep{Vickrey1961,Krishna2009}. Finally, the Keynesian \textbf{beauty-contest} (often operationalized as a \textbf{$p$-beauty contest} guessing game), a typically multi-player, simultaneous-move setting that is often repeated over rounds to probe higher-order belief reasoning and convergence \citep{Keynes1936,Nagel1995,DuffyNagel1997}.

%% file: latex/decision-makeing-data.tex
\begin{table*}
\centering
\resizebox{0.8\textwidth}{!}{%
\begin{tabular}{l ccccccc}
\toprule
& \textbf{Prisoner's} & \textbf{Stag} & \textbf{Escalation} & \textbf{Trust} & \textbf{Ultimatum} & \textbf{Sealed-Bid} & \textbf{Beauty} \\
& \textbf{Dilemma} & \textbf{Hunt} & \textbf{Game} & \textbf{Game} & \textbf{Game} & \textbf{Auction} & \textbf{Contest} \\
\midrule

\textbf{StarCraft} 
& - 
& - 
& 1042\textsuperscript{$\circlearrowright \diamondsuit$} 
& - 
& - 
& 1187\textsuperscript{$\circlearrowright \diamondsuit$} 
& - \\

\addlinespace
\textbf{Diplomacy} 
& 1264\textsuperscript{$\circlearrowright \diamondsuit \bullet$} 
& 1129\textsuperscript{$\circlearrowright \diamondsuit \bullet$} 
& 983\textsuperscript{$\circlearrowright \diamondsuit \bullet$} 
& 1076\textsuperscript{$\circlearrowright \diamondsuit \bullet$} 
& 1211\textsuperscript{$\circlearrowright \diamondsuit \bullet$} 
& 895\textsuperscript{$\circlearrowright \diamondsuit \bullet$} 
& 1098\textsuperscript{$\circlearrowright \diamondsuit \bullet$} \\

\addlinespace
\textbf{Personal Hub} 
& 912\textsuperscript{$\bullet$} 
& 847\textsuperscript{$\bullet$} 
& 1236\textsuperscript{$\circlearrowright \bullet$} 
& 1015\textsuperscript{$\circlearrowright \bullet$} 
& 1103\textsuperscript{$\circlearrowright \bullet$} 
& 1288\textsuperscript{$\circlearrowright \bullet$} 
& 969\textsuperscript{$\circlearrowright \bullet$} \\

\bottomrule
\end{tabular}
}
\caption{Data distribution across games and source. (\textit{Legend:} $\circlearrowright$ Multi-Turn; $\diamondsuit$ Multi-Modal; $\bullet$ Multi-Agent)}
\label{tab:game_attributes}

\end{table*}

Our dataset is constructed as a cross-product of \emph{game-theoretic decision templates} and heterogeneous \emph{data sources}. Templates fix the strategic form—actions, information structure, and payoff ordering—while source-specific pipelines generate labeled instances without altering the underlying mechanics. We use three sources: (i) No-Press Diplomacy episode mining, (ii) StarCraft~II (SC2) macro-decision extraction, and (iii) persona-conditioned scenario synthesis with real-world occupations.

\subsection{Diplomacy episodes}

We extract decision-making episodes from No-Press Diplomacy, a complex seven-player environment defined by cooperation and strategic depth without dialogue. Using rule-based detectors, we identify and filter localized interactions from match logs, spanning bi-player, multi-player, and multi-turn scenarios. These episodes are then converted via GPT-4.1-mini into natural-language narratives that preserve game-theoretic constraints and action sets while preventing information leakage.

\subsection{SC2 macro decisions}

To complement social-strategic interactions, we cast SC2 macro-management decisions—repeated resource allocations under partial observability—into Escalation Game and Non-Return Sealed-Bid templates. Using the MSC dataset~\citep{wu2017msc}, we scan replay trajectories for short windows matching these patterns via source-specific heuristics. Each window is rendered as a structured instance with scenario description, action choices, and template labels, aligned with match outcomes.

\subsection{Occupation-grounded synthesis}

We synthesize real-world scenarios to decouple strategic form from game-specific contexts. Job titles are sampled from PersonaHub and deduplicated via hashing and embeddings. Given a target template and participant jobs, GPT-4.1-mini generates job-consistent scenarios under an iterative generate–verify loop that enforces narrative coherence, incentive consistency, and payoff plausibility.

Across all sources, two authors annotate 80 random samples per template, achieving over 85\% agreement. To reduce lexical confounds in emotion interventions, we neutralize emotion-related wording in options using WordNet-Affect~\citep{strapparava-valitutti-2004-wordnet} and constrained rewriting. More details are in Appendix~\ref{app:synthesis_details}

%% file: latex/activation_steering.tex
\subsection{Emotion Eliciting Data Collection}
\label{subsec:emotion_text_collection}
\textbf{Emotion-eliciting Text Collection.} We source textual stimuli from the \textsc{crowd-enVENT} corpus, which contains crowd-written event descriptions annotated with emotion and appraisal labels, along with validation for annotation reliability \citep{troiano2023dimensional}. To ensure clear elicitation, we retain only validation instances with high inter-annotator agreement and restrict labels to six basic emotions: anger, joy, fear, disgust, sadness, and surprise. Because the original scenarios vary in surface form, we rewrite each retained item into a clean, syntactically uniform form using GPT-5.2.

\textbf{Emotion-eliciting Image Generation}. For each text scenario, we generate a paired visual stimulus with a pipeline designed to preserve viewpoint consistency, emotional salience, and selection stability. We prompt DALLE-3 to produce first-person images that depict witnessing the scenario, encouraging more subjective and affect-centered scenes. We then use majority voting with GPT-5.2 to select the best candidate image. More details are provided in Appendix~\ref{app:Emotion Eliciting Dataset}.

\textbf{Confounding Analysis}. To test whether stimuli could be classified using superficial stylistic cues, we examined eight shallow syntactic features: ratios of verbs, adjectives, adverbs, pronouns, determiners, adpositions, and coordinating conjunctions. A random forest classifier with 5-fold cross-validation achieved 0.274 accuracy, above the 0.167 chance level for six-way classification but still low overall. This suggests that only limited style-level confounds remain and that the stimuli are not reliably distinguishable from syntax alone. More details are in Appendix~\ref{app:Lexical and Syntactic Confounder}.

\subsection{Activation Steering}
\label{sec:activation_steering}

We adopt activation steering, which modulates model behaviour by adding a learned direction vector to intermediate activations during the forward pass, without updating model parameters \citep{rimsky2024caa,zou2023repe,zhang2025featureguided}. 

\paragraph{Obtaining emotion steering directions.}
\label{par:obtain_emotion_directions}
For each target emotion $e$, we construct a contrastive split with positives $\mathcal{D}_e$ and negatives $\mathcal{D}_{\neg e}$, where $\mathcal{D}_{\neg e}$ contains all samples labeled with other emotions. At decoder layer $\ell$, we extract the final hidden state $\mathbf{h}^{(\ell)}_{\text{last}}(\mathbf{x}) \in \mathbb{R}^{d}$. We then compute the target and non-target centroids,
\begin{equation}
\boldsymbol{\mu}^{(\ell)}_{e}
=
\frac{1}{|\mathcal{D}_{e}|}\sum_{\mathbf{x}\in\mathcal{D}_{e}} \mathbf{h}^{(\ell)}_{\text{last}}(\mathbf{x}),
\qquad
\boldsymbol{\mu}^{(\ell)}_{\neg e}
=
\frac{1}{|\mathcal{D}_{\neg e}|}\sum_{\mathbf{x}\in\mathcal{D}_{\neg e}} \mathbf{h}^{(\ell)}_{\text{last}}(\mathbf{x}),
\label{eq:mu_not_e}
\end{equation}
and define the contrast vector
\begin{equation}
\mathbf{c}^{(\ell,e)} = \boldsymbol{\mu}^{(\ell)}_{e} - \boldsymbol{\mu}^{(\ell)}_{\neg e}.
\label{eq:contrast_vector}
\end{equation}
Collecting mean-centered contrast vectors for emotion $e$ at layer $\ell$ yields a matrix $\mathbf{C}^{(\ell,e)} \in \mathbb{R}^{|\mathcal{D}_e|\times d}$. We apply principal component analysis (PCA) to $\mathbf{C}^{(\ell,e)}$ and use the first principal component as the steering direction:
\begin{equation}
\mathbf{s}^{(\ell,e)} \leftarrow \mathrm{PCA}_1\!\left(\mathbf{C}^{(\ell,e)}\right).
\label{eq:pca_direction}
\end{equation}
This produces layer-specific steering vectors that are directly compatible with residual-stream interventions across model families \citep{rimsky2024caa,zou2023repe}.

\paragraph{Steering the residual stream.}
Prior work suggests that emotion representations are concentrated in middle layers, so we select the middle third of decoder layers as control layers \citep{tigges2024linear, lee-etal-2025-large}. At each control layer, we intervene on the post-layer residual stream $\mathbf{h}^{(\ell)}_t \in \mathbb{R}^{d}$ at all token positions. Given steering vector $\mathbf{s}^{(\ell,e)}$, we add
\begin{equation}
\tilde{\mathbf{h}}^{(\ell)}_t = \mathbf{h}^{(\ell)}_t + \alpha \mathbf{s}^{(\ell,e)},
\label{eq:activation_addition}
\end{equation}
where $\alpha$ is the steering strength. We evaluate $\alpha \in \{0.6, 0.8, 1.0, 1.5\}$ and report average performance. The modified residual stream is then passed to subsequent layers, enabling a simple architecture-agnostic intervention that does not depend on attention-head structure, MLP decomposition, or gradient access \citep{vaswani2017attention,elhage2021framework}.

\subsection{Steering Validation}


\begin{wraptable}{r}{0.5\linewidth}
    
\centering
\resizebox{\linewidth}{!}{%
\begin{tabular}{lrrrrrrrrrr}
\toprule
& $\alpha$@1 & $\alpha$@4 & $\alpha$@8 & $\alpha$@10 & $\alpha$@15 & $\alpha$@20 & $\alpha$@40 & $\alpha$@80 \\ \midrule
anger     & 20/88 & 23/88 & 14/88 & 15/88 & 16/88 & 23/88 & 25/88 & 29/88 \\
happiness & 3/88  & 16/88 & 23/88 & 24/88 & 28/88 & 29/88 & 24/88 & 19/88 \\
sadness   & 0/88  & 8/88  & 10/88 & 15/88 & 13/88 & 15/88 & 19/88 & 17/88 \\
fear      & 0/88  & 0/88  & 0/88  & 0/88  & 1/88  & 2/88  & 7/88  & 4/88  \\
disgust   & 0/88  & 0/88  & 0/88  & 1/88  & 3/88  & 3/88  & 7/88  & 5/88  \\
surprise  & 0/88  & 0/88  & 3/88  & 1/88  & 4/88  & 4/88  & 4/88  & 6/88  \\ \bottomrule
\end{tabular}%
}
\caption{Fraction of layers with positive margin across all Qwen2.5 layers. Margin is defined as the token log-probability gap between the target emotion label and the strongest competing non-target label.}
\label{tab:self-report-emotion}
\end{wraptable}

To verify that our activation-steering method induces the intended affective state, we ran a separate self-report control experiment. We built a self-report benchmarkin which the model is asked to report its current emotional state by selecting one label from a set of seven options: six emotions and neutral under 10 different emotion-report prompts. We shuffled option orders to avoid position bias.  We computed the margin token log-probabilities between the target label and the strongest competing non-target label. Each time we only steer one layer , which means we can test larger $\alpha$. We sweep a wide range of $\alpha$ over all layers of Qwen2.5 series models.
Table~\ref{tab:self-report-emotion} shows  the ratio of layers with positive margin at certain intensities. We noticed an increasing trend of positive margin ratio across $\alpha$. 

%% file: latex/tab_emo_behav_alignment.tex
\begin{table*}[t]
\centering
\scriptsize
\setlength{\tabcolsep}{2.1pt}
\renewcommand{\arraystretch}{1.2}
\resizebox{\textwidth}{!}{%
\begin{tabular}{lccccccccccc}
\toprule
Model & 
\shortstack{Prisoners'\\Dilemma\\ \tiny NDM (w./ w.o.CoT) \\ \tiny NAD (w./ w.o.CoT)} & 
\shortstack{Stag\\Hunt\\ \tiny NDM (w./ w.o.CoT) \\ \tiny NAD (w./ w.o.CoT)} & 
\shortstack{Escalation\\Game\\ \tiny NDM (w./ w.o.CoT) \\ \tiny NAD (w./ w.o.CoT)} & 
\shortstack{Trust\\Trustor\\ \tiny NDM (w./ w.o.CoT) \\ \tiny NAD (w./ w.o.CoT)} & 
\shortstack{Trust\\Trustee\\ \tiny NDM (w./ w.o.CoT) \\ \tiny NAD (w./ w.o.CoT)} & 
\shortstack{Ultimatum\\Proposer\\ \tiny NDM (w./ w.o.CoT) \\ \tiny NAD (w./ w.o.CoT)} & 
\shortstack{Ultimatum\\Responder\\ \tiny NDM (w./ w.o.CoT) \\ \tiny NAD (w./ w.o.CoT)} & 
\shortstack{Beauty\\Contest\\ \tiny NDM (w./ w.o.CoT) \\ \tiny NAD (w./ w.o.CoT)} & 
\shortstack{Sealed\\Auction\\ \tiny NDM (w./ w.o.CoT) \\ \tiny NAD (w./ w.o.CoT)} & 
\shortstack{Mean\_Emotions\\ \tiny NDM (w./ w.o.CoT) \\ \tiny NAD (w./ w.o.CoT)} &
\shortstack{Mean\_Random\\ \tiny NDM (w./ w.o.CoT) \\ \tiny NAD (w./ w.o.CoT)} \\
\midrule
Llama-3.2-Instruct-1B & \shortstack{0.425 / 0.390 \\ 0.032 / 0.012} & \shortstack{0.407 / 0.359 \\ 0.023 / 0.042} & \shortstack{0.449 / 0.286 \\ 0.067 / -0.008} & \shortstack{0.399 / 0.364 \\ 0.050 / 0.031} & \shortstack{0.298 / 0.261 \\ 0.097 / 0.070} & \shortstack{0.372 / 0.311 \\ 0.018 / -0.006} & \shortstack{0.453 / 0.422 \\ 0.329 / 0.246} & \shortstack{0.368 / 0.364 \\ 0.008 / 0.018} & \shortstack{0.370 / 0.360 \\ -0.052 / -0.050} & \shortstack{0.393 / 0.346 \\ 0.063 / 0.039} & \shortstack{0.401 / 0.349 \\ 0.006 / -0.004} \\
\cmidrule(lr){2-12}
Llama-3.2-Instruct-3B & \shortstack{0.160 / 0.114 \\ 0.022 / 0.041} & \shortstack{0.119 / 0.064 \\ 0.062 / 0.038} & \shortstack{0.390 / 0.278 \\ 0.010 / -0.001} & \shortstack{0.318 / 0.226 \\ 0.205 / 0.123} & \shortstack{0.329 / 0.270 \\ 0.188 / 0.149} & \shortstack{0.331 / 0.280 \\ -0.018 / -0.021} & \shortstack{0.281 / 0.378 \\ 0.070 / 0.309} & \shortstack{0.297 / 0.206 \\ 0.008 / 0.006} & \shortstack{0.322 / 0.230 \\ 0.056 / 0.021} & \shortstack{0.283 / 0.227 \\ 0.067 / 0.074} & \shortstack{0.291 / 0.238 \\ 0.004 / 0.001} \\
\cmidrule(lr){2-12}

Phi-3.5-instruct-4B & \shortstack{0.042 / 0.291 \\ 0.008 / 0.046} & \shortstack{0.043 / 0.235 \\ 0.009 / 0.042} & \shortstack{0.168 / 0.315 \\ -0.003 / 0.004} & \shortstack{0.067 / 0.079 \\ 0.013 / 0.006} & \shortstack{0.183 / 0.183 \\ 0.009 / 0.010} & \shortstack{0.157 / 0.333 \\ 0.005 / 0.011} & \shortstack{0.223 / 0.323 \\ 0.110 / 0.208} & \shortstack{0.136 / 0.142 \\ 0.002 / 0.001} & \shortstack{0.160 / 0.168 \\ -0.006 / 0.000} & \shortstack{0.131 / 0.230 \\ 0.016 / 0.037} & \shortstack{0.139 / 0.225 \\ 0.003 / -0.002} \\
\cmidrule(lr){2-12}
Phi-4-instruct-4B & \shortstack{0.166 / 0.040 \\ 0.037 / 0.018} & \shortstack{0.213 / 0.085 \\ 0.076 / 0.064} & \shortstack{0.222 / 0.177 \\ -0.008 / 0.006} & \shortstack{0.195 / 0.064 \\ 0.074 / 0.043} & \shortstack{0.209 / 0.237 \\ 0.070 / 0.144} & \shortstack{0.225 / 0.114 \\ 0.016 / 0.011} & \shortstack{0.194 / 0.394 \\ 0.038 / 0.343} & \shortstack{0.152 / 0.098 \\ 0.008 / 0.007} & \shortstack{0.188 / 0.131 \\ 0.004 / 0.029} & \shortstack{0.196 / 0.149 \\ 0.035 / 0.074} & \shortstack{0.202 / 0.156 \\ 0.002 / -0.003} \\
\cmidrule(lr){2-12}

Qwen2.5-Instruct-0.5B & \shortstack{0.359 / 0.449 \\ 0.108 / 0.073} & \shortstack{0.259 / 0.314 \\ 0.074 / 0.062} & \shortstack{0.424 / 0.416 \\ -0.014 / -0.005} & \shortstack{0.338 / 0.354 \\ 0.046 / 0.036} & \shortstack{0.357 / 0.337 \\ 0.040 / 0.038} & \shortstack{0.427 / 0.386 \\ -0.013 / 0.002} & \shortstack{0.472 / 0.432 \\ 0.320 / 0.163} & \shortstack{0.361 / 0.334 \\ -0.007 / 0.012} & \shortstack{0.389 / 0.339 \\ -0.002 / -0.031} & \shortstack{0.376 / 0.373 \\ 0.061 / 0.039} & \shortstack{0.384 / 0.369 \\ 0.008 / 0.002} \\
\cmidrule(lr){2-12}
Qwen2.5-Instruct-1.5B & \shortstack{0.156 / 0.249 \\ 0.012 / 0.080} & \shortstack{0.174 / 0.232 \\ 0.030 / 0.053} & \shortstack{0.298 / 0.338 \\ 0.005 / 0.003} & \shortstack{0.254 / 0.290 \\ 0.069 / 0.045} & \shortstack{0.292 / 0.282 \\ 0.056 / 0.047} & \shortstack{0.415 / 0.229 \\ -0.027 / -0.024} & \shortstack{0.471 / 0.430 \\ 0.205 / 0.157} & \shortstack{0.298 / 0.208 \\ 0.005 / 0.004} & \shortstack{0.306 / 0.243 \\ -0.016 / 0.032} & \shortstack{0.296 / 0.278 \\ 0.038 / 0.044} & \shortstack{0.304 / 0.281 \\ 0.001 / -0.005} \\
\cmidrule(lr){2-12}
Qwen2.5-Instruct-3B & \shortstack{0.200 / 0.193 \\ 0.017 / 0.072} & \shortstack{0.294 / 0.113 \\ 0.063 / 0.045} & \shortstack{0.312 / 0.068 \\ 0.014 / 0.013} & \shortstack{0.243 / 0.263 \\ 0.019 / 0.018} & \shortstack{0.266 / 0.280 \\ 0.040 / 0.060} & \shortstack{0.391 / 0.326 \\ -0.018 / -0.021} & \shortstack{0.231 / 0.235 \\ 0.041 / 0.132} & \shortstack{0.353 / 0.159 \\ 0.005 / -0.005} & \shortstack{0.383 / 0.190 \\ 0.008 / 0.001} & \shortstack{0.297 / 0.203 \\ 0.021 / 0.035} & \shortstack{0.305 / 0.211 \\ -0.004 / 0.003} \\
\cmidrule(lr){2-12}

Qwen3-0.6B & \shortstack{0.154 / 0.151 \\ 0.013 / 0.031} & \shortstack{0.165 / 0.134 \\ 0.027 / 0.048} & \shortstack{0.283 / 0.197 \\ -0.006 / -0.010} & \shortstack{0.191 / 0.159 \\ 0.050 / 0.008} & \shortstack{0.183 / 0.173 \\ 0.059 / 0.025} & \shortstack{0.172 / 0.199 \\ 0.014 / -0.003} & \shortstack{0.232 / 0.246 \\ 0.164 / 0.215} & \shortstack{0.211 / 0.156 \\ 0.003 / 0.008} & \shortstack{0.171 / 0.092 \\ -0.007 / 0.007} & \shortstack{0.196 / 0.167 \\ 0.035 / 0.037} & \shortstack{0.201 / 0.171 \\ 0.002 / -0.004} \\
\cmidrule(lr){2-12}

Qwen3-1.7B & \shortstack{0.086 / 0.143 \\ -0.006 / 0.009} & \shortstack{0.014 / 0.012 \\ 0.003 / 0.000} & \shortstack{0.253 / 0.242 \\ 0.004 / 0.006} & \shortstack{0.156 / 0.150 \\ 0.003 / -0.019} & \shortstack{0.165 / 0.205 \\ -0.001 / -0.003} & \shortstack{0.307 / 0.337 \\ 0.010 / -0.007} & \shortstack{0.061 / 0.093 \\ 0.009 / 0.025} & \shortstack{0.205 / 0.122 \\ -0.007 / 0.001} & \shortstack{0.137 / 0.111 \\ -0.011 / -0.013} & \shortstack{0.154 / 0.157 \\ 0.000 / -0.000} & \shortstack{0.161 / 0.154 \\ 0.001 / -0.002} \\
\cmidrule(lr){2-12}

Qwen3-4B & \shortstack{0.189 / 0.099 \\ 0.026 / 0.016} & \shortstack{0.199 / 0.015 \\ 0.024 / 0.007} & \shortstack{0.340 / 0.106 \\ 0.020 / 0.022} & \shortstack{0.239 / 0.163 \\ -0.006 / -0.005} & \shortstack{0.177 / 0.086 \\ -0.008 / 0.011} & \shortstack{0.248 / 0.334 \\ 0.019 / 0.012} & \shortstack{0.260 / 0.244 \\ 0.050 / 0.080} & \shortstack{0.235 / 0.181 \\ -0.001 / -0.006} & \shortstack{0.263 / 0.193 \\ -0.002 / 0.022} & \shortstack{0.239 / 0.158 \\ 0.014 / 0.018} & \shortstack{0.245 / 0.162 \\ 0.003 / -0.001} \\
\cmidrule(lr){2-12}

Gemma-3-it-270M & \shortstack{0.214 / 0.231 \\ 0.007 / 0.015} & \shortstack{0.226 / 0.256 \\ -0.007 / -0.003} & \shortstack{0.289 / 0.194 \\ 0.002 / 0.002} & \shortstack{0.317 / 0.400 \\ 0.019 / -0.006} & \shortstack{0.330 / 0.393 \\ -0.013 / -0.000} & \shortstack{0.334 / 0.381 \\ 0.013 / 0.006} & \shortstack{0.189 / 0.022 \\ -0.017 / 0.004} & \shortstack{0.318 / 0.395 \\ -0.006 / 0.005} & \shortstack{0.331 / 0.425 \\ 0.002 / 0.001} & \shortstack{0.283 / 0.300 \\ -0.000 / 0.003} & \shortstack{0.289 / 0.303 \\ 0.001 / -0.002} \\
\cmidrule(lr){2-12}

Gemma-3-it-1B & \shortstack{0.215 / 0.143 \\ -0.004 / -0.006} & \shortstack{0.077 / 0.042 \\ -0.001 / 0.004} & \shortstack{0.239 / 0.075 \\ 0.010 / -0.005} & \shortstack{0.153 / 0.248 \\ 0.007 / 0.012} & \shortstack{0.152 / 0.212 \\ 0.002 / 0.003} & \shortstack{0.370 / 0.387 \\ -0.003 / -0.002} & \shortstack{0.337 / 0.287 \\ -0.038 / 0.004} & \shortstack{0.233 / 0.212 \\ -0.001 / 0.002} & \shortstack{0.211 / 0.185 \\ -0.005 / 0.017} & \shortstack{0.221 / 0.199 \\ -0.004 / 0.003} & \shortstack{0.227 / 0.203 \\ 0.002 / -0.001} \\
\cmidrule(lr){2-12}

Gemma-3-it-4B & \shortstack{0.188 / 0.386 \\ -0.006 / 0.009} & \shortstack{0.154 / 0.159 \\ -0.001 / -0.002} & \shortstack{0.324 / 0.368 \\ -0.000 / -0.009} & \shortstack{0.239 / 0.115 \\ 0.007 / 0.001} & \shortstack{0.204 / 0.132 \\ 0.005 / 0.005} & \shortstack{0.378 / 0.274 \\ 0.011 / -0.003} & \shortstack{0.259 / 0.284 \\ -0.004 / 0.015} & \shortstack{0.232 / 0.155 \\ 0.002 / 0.000} & \shortstack{0.236 / 0.178 \\ -0.012 / -0.005} & \shortstack{0.246 / 0.228 \\ 0.000 / 0.001} & \shortstack{0.252 / 0.231 \\ 0.001 / -0.002} \\

\midrule
Zamba2-Instruct-1.2B & \shortstack{0.359 / 0.301 \\ 0.097 / 0.089} & \shortstack{0.320 / 0.180 \\ 0.073 / 0.053} & \shortstack{0.366 / 0.141 \\ 0.014 / 0.009} & \shortstack{0.350 / 0.188 \\ 0.012 / 0.001} & \shortstack{0.291 / 0.199 \\ 0.006 / -0.003} & \shortstack{0.348 / 0.284 \\ 0.004 / 0.028} & \shortstack{0.455 / 0.317 \\ 0.253 / 0.247} & \shortstack{0.364 / 0.348 \\ 0.001 / 0.006} & \shortstack{0.432 / 0.426 \\ -0.004 / -0.030} & \shortstack{0.365 / 0.265 \\ 0.051 / 0.044} & \shortstack{0.371 / 0.271 \\ 0.005 / -0.003} \\
\cmidrule(lr){2-12}

Zamba2-Instruct-2.7B & \shortstack{0.252 / 0.228 \\ 0.015 / 0.028} & \shortstack{0.169 / 0.221 \\ 0.058 / 0.056} & \shortstack{0.397 / 0.057 \\ 0.001 / 0.004} & \shortstack{0.279 / 0.306 \\ 0.082 / 0.032} & \shortstack{0.270 / 0.252 \\ 0.080 / 0.021} & \shortstack{0.387 / 0.356 \\ -0.018 / 0.006} & \shortstack{0.481 / 0.464 \\ 0.338 / 0.332} & \shortstack{0.333 / 0.342 \\ -0.006 / -0.002} & \shortstack{0.390 / 0.374 \\ 0.010 / 0.004} & \shortstack{0.329 / 0.289 \\ 0.062 / 0.054} & \shortstack{0.336 / 0.294 \\ 0.003 / -0.002} \\
\cmidrule(lr){2-12}

Mamba2-1.3B & \shortstack{0.281 / 0.236 \\ 0.041 / 0.033} & \shortstack{0.198 / 0.154 \\ 0.036 / 0.028} & \shortstack{0.318 / 0.129 \\ 0.008 / 0.005} & \shortstack{0.267 / 0.201 \\ 0.031 / 0.018} & \shortstack{0.249 / 0.187 \\ 0.027 / 0.014} & \shortstack{0.301 / 0.248 \\ 0.006 / 0.011} & \shortstack{0.392 / 0.281 \\ 0.186 / 0.173} & \shortstack{0.287 / 0.264 \\ 0.002 / 0.004} & \shortstack{0.341 / 0.318 \\ 0.003 / -0.010} & \shortstack{0.293 / 0.224 \\ 0.038 / 0.031} & \shortstack{0.299 / 0.229 \\ 0.002 / -0.001} \\
\cmidrule(lr){2-12}

Mamba2-2.7B & \shortstack{0.334 / 0.287 \\ 0.022 / 0.027} & \shortstack{0.241 / 0.209 \\ 0.049 / 0.041} & \shortstack{0.372 / 0.103 \\ 0.006 / 0.007} & \shortstack{0.301 / 0.279 \\ 0.061 / 0.026} & \shortstack{0.293 / 0.261 \\ 0.054 / 0.020} & \shortstack{0.356 / 0.318 \\ -0.006 / 0.007} & \shortstack{0.444 / 0.421 \\ 0.281 / 0.269} & \shortstack{0.314 / 0.307 \\ -0.003 / -0.001} & \shortstack{0.368 / 0.351 \\ 0.008 / 0.003} & \shortstack{0.336 / 0.282 \\ 0.052 / 0.044} & \shortstack{0.343 / 0.286 \\ 0.004 / -0.002} \\

\midrule
Phi-3.5-vision-4B & \shortstack{0.070 / 0.245 \\ 0.009 / 0.013} & \shortstack{0.031 / 0.126 \\ 0.006 / 0.020} & \shortstack{0.208 / 0.288 \\ 0.005 / 0.007} & \shortstack{0.057 / 0.086 \\ 0.002 / 0.012} & \shortstack{0.161 / 0.171 \\ -0.011 / -0.001} & \shortstack{0.299 / 0.271 \\ 0.004 / -0.002} & \shortstack{0.141 / 0.087 \\ 0.018 / 0.028} & \shortstack{0.153 / 0.142 \\ -0.007 / -0.002} & \shortstack{0.127 / 0.179 \\ 0.020 / 0.001} & \shortstack{0.139 / 0.177 \\ 0.005 / 0.009} & \shortstack{0.145 / 0.181 \\ 0.002 / -0.001} \\
\cmidrule(lr){2-12}
Phi-4-multimodal-4B & \shortstack{0.058 / 0.025 \\ 0.008 / -0.001} & \shortstack{0.115 / 0.057 \\ 0.014 / -0.009} & \shortstack{0.211 / 0.124 \\ -0.007 / 0.008} & \shortstack{0.099 / 0.014 \\ -0.005 / 0.000} & \shortstack{0.162 / 0.078 \\ 0.005 / -0.008} & \shortstack{0.090 / 0.129 \\ 0.022 / 0.008} & \shortstack{0.190 / 0.125 \\ 0.033 / 0.022} & \shortstack{0.049 / 0.094 \\ 0.003 / -0.003} & \shortstack{0.192 / 0.123 \\ 0.092 / -0.013} & \shortstack{0.129 / 0.085 \\ 0.018 / 0.001} & \shortstack{0.135 / 0.089 \\ 0.004 / -0.002} \\
\cmidrule(lr){2-12}
Qwen3-VL-Instruct-2B & \shortstack{0.106 / 0.233 \\ 0.005 / 0.019} & \shortstack{0.104 / 0.059 \\ 0.010 / -0.003} & \shortstack{0.333 / 0.219 \\ 0.013 / 0.003} & \shortstack{0.195 / 0.163 \\ 0.009 / -0.004} & \shortstack{0.166 / 0.210 \\ -0.004 / -0.009} & \shortstack{0.205 / 0.330 \\ -0.004 / 0.013} & \shortstack{0.011 / 0.074 \\ 0.002 / 0.016} & \shortstack{0.000 / 0.158 \\ 0.000 / 0.002} & \shortstack{0.011 / 0.066 \\ -0.011 / 0.001} & \shortstack{0.126 / 0.168 \\ 0.002 / 0.004} & \shortstack{0.132 / 0.172 \\ 0.001 / -0.002} \\
\cmidrule(lr){2-12}
Qwen3-VL-Instruct-4B & \shortstack{0.167 / 0.072 \\ 0.009 / 0.023} & \shortstack{0.231 / 0.046 \\ 0.048 / 0.022} & \shortstack{0.291 / 0.213 \\ 0.002 / 0.007} & \shortstack{0.255 / 0.095 \\ 0.022 / 0.002} & \shortstack{0.149 / 0.097 \\ 0.045 / 0.029} & \shortstack{0.258 / 0.209 \\ 0.024 / 0.003} & \shortstack{0.059 / 0.038 \\ 0.033 / 0.025} & \shortstack{0.241 / 0.142 \\ -0.002 / -0.007} & \shortstack{0.312 / 0.171 \\ 0.035 / -0.003} & \shortstack{0.218 / 0.120 \\ 0.024 / 0.011} & \shortstack{0.224 / 0.126 \\ 0.003 / -0.001} \\
\cmidrule(lr){2-12}

InternVL3-1B & \shortstack{0.251 / 0.260 \\ 0.032 / 0.012} & \shortstack{0.097 / 0.121 \\ 0.016 / -0.001} & \shortstack{0.369 / 0.263 \\ -0.015 / -0.006} & \shortstack{0.311 / 0.244 \\ 0.088 / 0.009} & \shortstack{0.308 / 0.271 \\ 0.066 / 0.002} & \shortstack{0.422 / 0.341 \\ 0.060 / 0.003} & \shortstack{0.486 / 0.156 \\ 0.381 / 0.052} & \shortstack{0.321 / 0.316 \\ 0.006 / 0.012} & \shortstack{0.332 / 0.286 \\ 0.009 / 0.019} & \shortstack{0.322 / 0.251 \\ 0.071 / 0.011} & \shortstack{0.328 / 0.255 \\ 0.004 / -0.003} \\
\cmidrule(lr){2-12}
InternVL3-2B & \shortstack{0.286 / 0.298 \\ 0.003 / 0.040} & \shortstack{0.228 / 0.336 \\ 0.036 / 0.014} & \shortstack{0.347 / 0.311 \\ 0.009 / -0.001} & \shortstack{0.285 / 0.303 \\ -0.003 / 0.021} & \shortstack{0.310 / 0.239 \\ 0.037 / 0.030} & \shortstack{0.390 / 0.231 \\ -0.025 / 0.003} & \shortstack{0.383 / 0.464 \\ 0.099 / 0.086} & \shortstack{0.335 / 0.193 \\ 0.003 / 0.001} & \shortstack{0.309 / 0.235 \\ 0.011 / 0.040} & \shortstack{0.319 / 0.290 \\ 0.019 / 0.026} & \shortstack{0.325 / 0.294 \\ 0.002 / -0.001} \\

\bottomrule
\end{tabular}
}

\caption{Overall results on normalized drift magnitude  and  normalized aligned drift with a comparision between raw output and CoT mode. `Mean\_Random' column shows the mean results under a random steering vector. `Mean\_Emotions' column shows mean results over all emotion steerings.}
\label{tab:combined_results}
\end{table*}

%% file: latex/fig_heatmap_main.tex
\begin{figure*}[h]
    \centering
    \includegraphics[width=1\linewidth]{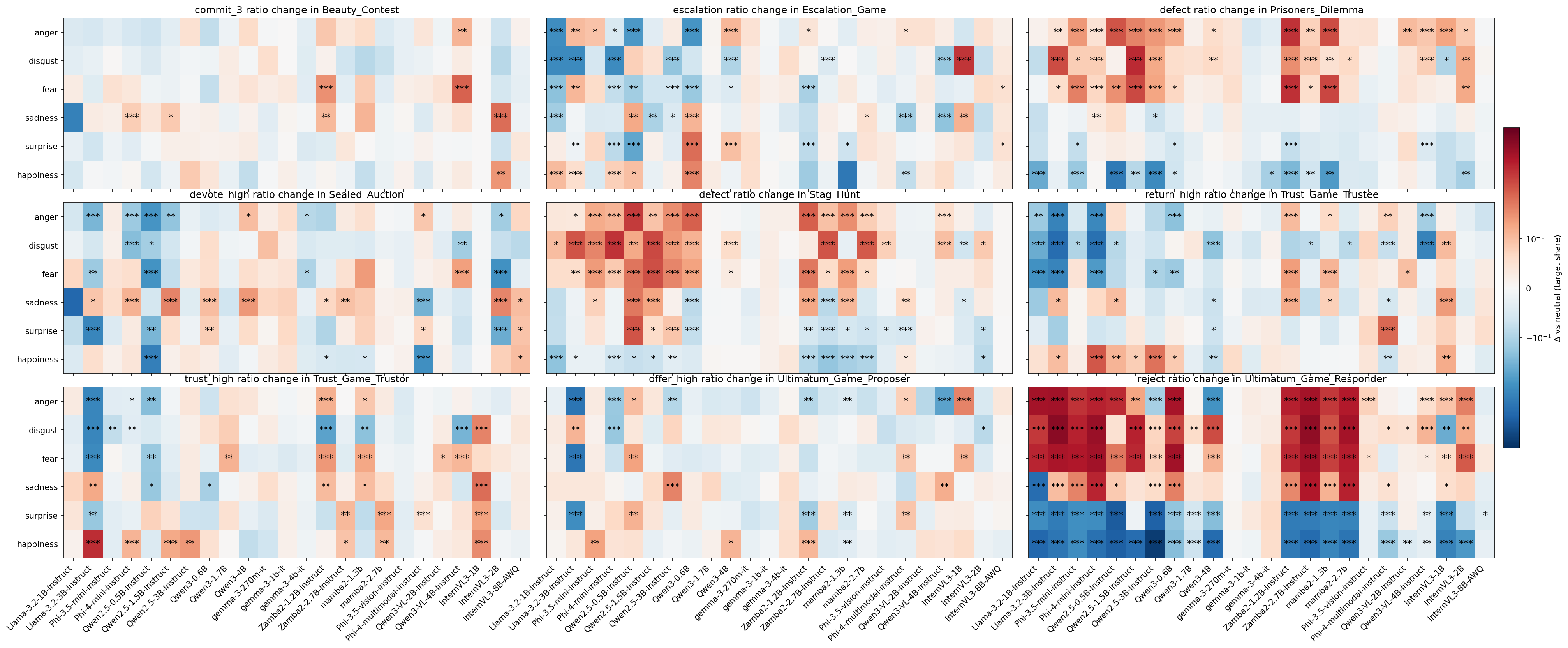}
    \caption{Emotion-eliciting decision shift for different model - game combination. We can observe a obvious difference between positive emotions and negative emotions. Annotations indicate FDR-adjusted significance thresholds (*  (q<0.05), ** (q<0.01), *** (q<0.001))}
    \label{fig:heatmap}
\end{figure*}

%% file: latex/results.tex

\label{sec:results}


To characterize emotion steering in decision making, we evaluate it along two complementary dimensions: Normalized Drift Magnitude (NDM) and Normalized Aligned Drift (NAD). NDM answers how far emotion steering moves a model away from its neutral response.
\begin{equation}
{\small
\begin{aligned}
\mathrm{NDM}&=\mathbb{E}[|y_i^{(e)}-y_i^{(0)}|/R_i], \qquad
R_i=y_i^{\max}-y_i^{\min},\\
\mathrm{NAD}&=\mathbb{E}[d_i^{\mathrm{human}}(y_i^{(e)}-y_i^{(0)})/R_i], \qquad
d_i^{\mathrm{human}}\in\{-1,0,+1\}.
\end{aligned}
}
\end{equation}

where \(y_i^{(0)}\) denotes the neutral decision, \(y_i^{(e)}\) the decision under emotion \(e\), and \(R_i\) the feasible response range of item \(i\). 

NAD assesses whether models exhibit human-like emotional response patterns . For each (game, emotion) situation , we build \(d_i^{\mathrm{human}}\) to represent literature-grounded expected direction derived from findings in experimental economics and psychology.

Positive NAD values indicate that emotion steering shifts behavior in the human-expected direction on average. More details on the construction of \(d_i^{\mathrm{human}}\) are provided in Appendix~\ref{app:Behavior shift alignment}. Unless otherwise specified, all reported results use greedy decoding without Chain-of-Thought (CoT) reasoning to improve reproducibility.

\subsection{Emotion as an unstable control knob}
\label{subsec:unstable_knob}

Table~\ref{tab:combined_results} reports NAD and NDM values over all model-game combination.  Overall, the results suggest that emotion steering can induce partially human-consistent directional changes, but these effects remain highly model- and task-dependent. As shown in Fig.~\ref{fig:heatmap}, statistical significance was assessed using an exact  McNemar test on paired  neutral-emotion  responses,   To control the false discovery  rate across the set of model-emotion comparisons within each heatmap,  raw p-values were adjusted using the Benjamini-Hochberg procedure. Heatmap annotations indicate significance based on the resulting FDR-adjusted q-values
Positive emotions such as happiness and surprise are more often associated with increased trust, sharing, or cooperation, whereas negative emotions such as anger and disgust are more often associated with avoidance, or punishment-oriented behavior \citep{lerner2015emotion}. This pattern appears across model architectures and modalities. For example in Ultimatum (Responder), where several text-only models show consistently positive alignment, including Qwen2.5-0.5B ($.320/0.163$), Zamba2-2.7B ($0.338 / 0.332$) with/without CoT.

At the same time, emotion is not a stable or model-invariant control knob for producing human-faithful directional changes. Some model series like Qwen2.5 show good alignment, while others  like Gemma changes decisions with average-level NDM but almost zero NAD. In some tasks, model behavior even reverses common human expectations. This is clearest in the Escalation Game, where several models show weak or negative alignment, including Qwen2.5-0.5B ($-0.014 / -0.005$), InternVL3-1B ($-0.015 / -0.006$) . Such reversals suggest that emotion directions in model representation space do not consistently map onto human appraisal-to-action mechanisms.

\input{latex/intensity_impacts}

\subsection{Emotion Effects Persist Under Sampling Decoding}

Although our main results disable sampling to reduce randomness, we repeated the analysis with stochastic decoding (\(Temperature=0.7\)) to test whether the conclusions changed. We focused on the Qwen2.5 series, which showed the highest NAD. The analysis covered three Qwen2.5 models, seven game-theory benchmarks, six emotions, and three intensity levels, yielding 378 emotion conditions. For each condition, we compared the emotion manipulation with a matched neutral baseline using a Cochran--Mantel--Haenszel test stratified by sampling repeat, with pooled change in target-behavior rate as the effect size. Results were broadly consistent: \(256/378\) conditions (\(67.7\%\)) were significant. Effect direction was also stable across repeats. Only \(71/378\) conditions (\(18.8\%\)) changed sign, and no condition showed significant Breslow--Day heterogeneity. The strongest effects were therefore highly reproducible under sampling.

\subsection{How reasoning interacts with emotion impacts}
\label{subsec:thinking_no_gain}
Many studies have shown that 'thinking-efforts' can enhance performance across diverse tasks. To investigate the influence of reasoning on emotional decision-making, we examine two distinct approaches: chain-of-thought (CoT) prompting in standard models and specialized reasoning models, such as Qwen3-Thinking, which are explicitly trained for advanced cognitive capabilities.

\begin{figure}[htbp]
    \centering
    \includegraphics[width=0.8\linewidth]{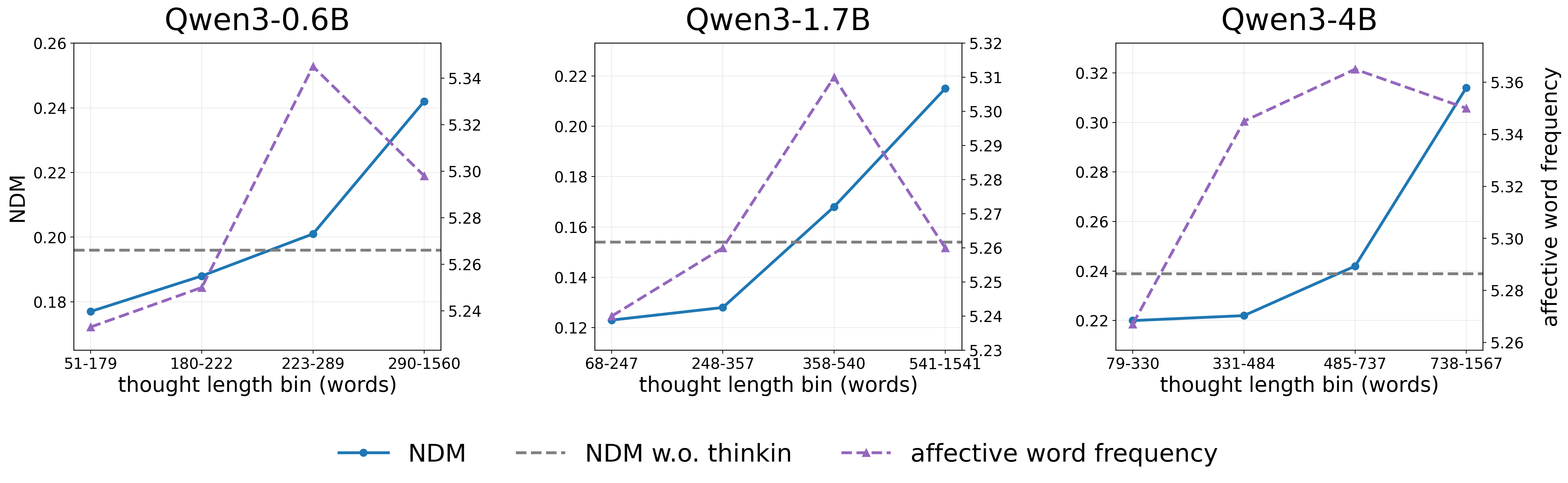}
\caption{The vulnerability of Qwen3 thinking mode increases with thought length and affective word frequency.}    
\label{fig:qwen3plots}
\end{figure}

As shown in Table~\ref{tab:combined_results} ,  18/23 (78\%) models show higher NDM with CoT , which aligned with results from  Qwen3-Thinking.  As shown in Fig~\ref{fig:qwen3plots}, across the Qwen3 family, enabling thinking mode does not reduce NDM to emotional manipulation and even makes agents more sensitive to emotion. Compared to the non-thinking condition, the thinking runs exhibit higher NDM. These results  indicate that longer internal deliberation does not “average out” affective perturbations but can amplify them. A plausible mechanism is accumulation: when an emotion-induced bias perturbs early intermediate representations, reasoning provides more opportunities for that bias to be reinforced. This interpretation is supported by the correlation between thinking length and choice change rate, which can be clearly observed in Qwen3-Thinking's reasoning process. We use WordNet-Affect to identify “affective words” for each emotion. As shown in Figure~\ref{fig:qwen3plots}, as mean thinking length increases, both the NDM and the “affective words” number increase together.  

To better understand which items are more affected by CoT process, we further conduct an item-level psychometric analysis to better understand CoT impacts. We use a Bayesian 2PL model for binary tasks and an ordered logistic model for ordered-response tasks to estimate neutral-response item difficulty and discrimination. Conditioned on emotion and game,  we run a  regression on independent variables of item difficulty and discrimination , and delta NDM , delta NAD set as  dependent variables respectively. In both models, the resulting coefficients were statistically reliable but substantively small. For example,  when delta NAD as dependent variable , item difficulty show beta = 0.0245, p < 1e-4, which is significant but with  low effect size. More details about IRT estimation, see Appendix~\ref{latxIRT_analy}

\section{Alleviate emotion effects by thought audits}
\label{sec:thoughts audit}
\begin{wraptable}{r}{0.4\textwidth}
\vspace{-10pt} 
\centering
\caption{ NDM is alleviated by thoughts audit, showing improvements in robustness}
\resizebox{0.4\textwidth}{!}{
\begin{tabular}{lcc}
\toprule
Models & \shortstack{w.o. audit\\NDM/NAD}& \shortstack{w. audit\\NDM/NAD} \\
\midrule
Llama-3.2-3B-Instruct & 0.283/0.056 & 0.264/0.058\\
Phi-4-mini-Instruct   & 0.196/0.035 & 0.179/0.034\\
Qwen2.5-3B-Instruct   & 0.297/0.021& 0.284/0.019 \\
Qwen3-4B              & 0.239/0.014 & 0.221/0.011 \\
Gemma-3-4B-it         & 0.246/0.000 & 0.237/0.000 \\
\bottomrule
\end{tabular}
}
\label{tab:alleviate_results}
\vspace{-10pt} 
\end{wraptable}

The observation from Qwen3 thinking mode experiments inspires us to use audit models reasoning process to alleviate the emotion effects.  We train a lightweight predictor on model's emotional CoT. We emply a simple TF--IDF with logistic regression method train on unigrams in cot $t$ under emotion $e$ and outputs $p_i^{(e)}=\Pr(y_i^{(e)}=1\mid t_i^{(e)})$. This simple predictor can achieve AUC=0.81. 

At inference, the predictor acts as a gatekeeper: we choose $\tau$ as threshold for high accuracy then flag items with $p_i^{(e)}\ge \tau$  and route only those to a second-turn reflection prompt; all others keep their original decisions. As shown in Table~\ref{tab:alleviate_results},  this simple method reduces NDM and does not change NAD too much. It  demonstrates that thoughts audit mitigate decision shift in all directions evenly.

\section{Conclusion}
\label{sec:conclusion}

We build a decision making dataset and conduct intensive experiments on it. Based on the results, we find that emotion steering yields pervasive decision vulnerability across model families and game settings, while offering only limited human-aligned directional control. The emotion-based behavior shift is not predictably transferable across models and, in some games, systematically departs from human-intuitive affective patterns.

\section*{Reproducibility Statement}
\label{sec:Reproducibility}

 The evaluation abstracts decisions into discrete actions under a fixed elicitation and steering protocol. Although the benchmark spans multiple canonical games, broader multi-agent and more naturalistic interactions may exhibit additional failure modes. Besides using AI for dataset generation, we use AI for the writing , coding and assist analysing experimental results for this paper. Although we checked all artifacts generated by AI , it is still possible to have unseen problems that will impact the results of our paper.

%% file: latex/intensity_impacts.tex
\subsection{Steering intensity amplifies emotional impact}
\label{subsec:intensity_impact}
To isolate the effect of steering strength, we compute the normalized  NDM across alphas
Table~\ref{tab:intensity_impact_model} shows that, for a large proportion of models, NDM increases with steering intensity. Text-only models such as Llama-3.2-1B, Phi-4-mini, and Qwen2.5-0.5B exhibit clear monotonic growth from $\alpha=0.6$ to $1.5$, while multimodal models such as InternVL3-1B and Qwen3-VL-4B demonstrate similar patterns. By contrast, some models remain stable across all intensity levels; for instance, the Gemma-3 models are nearly invariant to these interventions. These findings suggest that sensitivity to steering intensity is highly model-dependent. Nevertheless, among models that do exhibit substantial shifts in emotional behavior, the magnitude of change is positively associated with intensity.

\begin{table*}[h]
    \centering
    \scriptsize
    \setlength{\tabcolsep}{3pt}
    \renewcommand{\arraystretch}{1.12}
    \resizebox{\textwidth}{!}{%
$$\begin{tabular}{lccccccccccccccccccccc}
\toprule
Metric &
\shortstack{Llama-3.2\\1B} &
\shortstack{Llama-3.2\\3B} &
\shortstack{Phi\\3.5-mini} &
\shortstack{Phi\\4-mini} &
\shortstack{Qwen2.5\\0.5B} &
\shortstack{Qwen2.5\\1.5B} &
\shortstack{Qwen2.5\\3B} &
\shortstack{Qwen3\\0.6B} &
\shortstack{Qwen3\\1.7B} &
\shortstack{Qwen3\\4B} &
\shortstack{Gemma-3\\270M} &
\shortstack{Gemma-3\\1B} &
\shortstack{Gemma-3\\4B} &
\shortstack{Phi-3.5\\vision} &
\shortstack{Phi-4\\multi} &
\shortstack{Qwen3-VL\\2B} &
\shortstack{Qwen3-VL\\4B} &
\shortstack{Qwen3-VL\\4B-Thk} &
\shortstack{InternVL3\\1B} &
\shortstack{InternVL3\\2B} &
\shortstack{InternVL3\\8B} \\
\midrule
NDM($0.6$) & 0.3371 & 0.1915 & 0.2146 & 0.1182 & 0.3434 & 0.2597 & 0.1770 & 0.1417 & 0.1548 & 0.1496 & 0.3043 & 0.1986 & 0.2272 & 0.1743 & 0.0697 & 0.1638 & 0.1057 & 0.6667 & 0.2000 & 0.2842 & 0.0657 \\
NAD($0.6$) & 0.0067 & 0.0497 & 0.0219 & 0.0523 & 0.0307 & 0.0239 & 0.0267 & 0.0214 & -0.0010 & 0.0121 & 0.0028 & 0.0029 & 0.0008 & 0.0039 & -0.0011 & 0.0012 & 0.0041 & 0.0741 & 0.0113 & 0.0140 & -0.0023 \\
\midrule
NDM($0.8$) & 0.3485 & 0.2096 & 0.2224 & 0.1335 & 0.3584 & 0.2688 & 0.1900 & 0.1547 & 0.1560 & 0.1540 & 0.3041 & 0.1989 & 0.2275 & 0.1765 & 0.0779 & 0.1663 & 0.1143 & 0.5000 & 0.2289 & 0.2876 & 0.0679 \\
NAD($0.8$) & -0.0091 & 0.0635 & 0.0291 & 0.0632 & 0.0338 & 0.0344 & 0.0304 & 0.0293 & -0.0005 & 0.0152 & 0.0027 & 0.0033 & 0.0010 & 0.0064 & 0.0001 & 0.0030 & 0.0082 & 0.0556 & 0.0113 & 0.0211 & -0.0023 \\
\midrule
NDM($1.0$) & 0.3741 & 0.2270 & 0.2303 & 0.1491 & 0.3739 & 0.2774 & 0.2033 & 0.1668 & 0.1576 & 0.1571 & 0.3050 & 0.1996 & 0.2279 & 0.1760 & 0.0836 & 0.1671 & 0.1176 & 0.5000 & 0.2442 & 0.2895 & 0.0699 \\
NAD($1.0$) & -0.0203 & 0.0755 & 0.0374 & 0.0765 & 0.0414 & 0.0431 & 0.0362 & 0.0358 & -0.0001 & 0.0173 & 0.0027 & 0.0025 & 0.0013 & 0.0082 & -0.0006 & 0.0035 & 0.0104 & 0.0000 & 0.0106 & 0.0243 & -0.0012 \\
\midrule
NDM($1.5$) & 0.3784 & 0.2454 & 0.2381 & 0.1642 & 0.3886 & 0.2868 & 0.2161 & 0.1802 & 0.1586 & 0.1622 & 0.3042 & 0.1997 & 0.2281 & 0.1796 & 0.0931 & 0.1704 & 0.1289 & 0.2500 & 0.2799 & 0.2937 & 0.0722 \\
NAD($1.5$) & -0.0384 & 0.0902 & 0.0441 & 0.0862 & 0.0422 & 0.0545 & 0.0388 & 0.0444 & 0.0004 & 0.0209 & 0.0025 & 0.0035 & 0.0014 & 0.0111 & 0.0016 & 0.0059 & 0.0154 & 0.0000 & 0.0110 & 0.0333 & -0.0017 \\
\bottomrule
\end{tabular}$$    
    }
    \caption{Change of NDM and NAD across various $\alpha$. }
    \label{tab:intensity_impact_model}
    \end{table*}

%% file: appendix/decision_making_dataset_appendix.tex
\section{Source-Specific Pipelines and Methodology}
\label{app:pipelines_natural_language}

\subsection{Diplomacy Episode Mining}
\label{app:diplomacy_details}

The Diplomacy pipeline extracts strategic interactions from "No-Press" match logs by identifying specific temporal windows that align with game-theoretic templates.

\begin{itemize}
    \item \textbf{Preprocessing and Parsing:} For every match log in the corpus, the system first verifies data integrity. It then parses the sequence of phases and orders into a structured timeline ($\Pi$).
    \item \textbf{Window Identification:} For each strategic template $T$ (e.g., Trust, Ultimatum), the system scans the timeline to propose \textit{candidate windows} ($w$) where relevant interactions might occur.
    \item \textbf{Trigger and Validation:} A candidate window is promoted to a dataset record if it satisfies two conditions:
    \begin{enumerate}
        \item \textbf{Trigger Logic:} The specific board state and order sequence must meet the mathematical criteria defined by the template (e.g., an increase in exposure or a credible threat).
        \item \textbf{Participant Validation:} The system confirms that the actors involved are logically positioned to fulfill the roles required by the template.
    \end{enumerate}
    \item \textbf{Rendering:} Validated windows are rendered into a standardized JSONL format, creating the final episode dataset $\mathcal{D}_{dip}$.
\end{itemize}

\subsubsection{Template Trigger Definitions}
The \textsc{Trigger} function serves as a strategic filter. Key definitions include:
\begin{itemize}
    \item \textbf{Trust:} Requires a verified increase in one player's vulnerability to another (\textit{Exposure Increase}) combined with a future phase that allows for a reciprocal act (\textit{Reciprocation Opportunity}).
    \item \textbf{Ultimatum:} Requires a player to have a \textit{Credible Threat} (the ability to capture a center) and a \textit{Veto Fork}, where the responder's choice directly determines the success or failure of that threat.
    \item \textbf{Beauty Contest:} Triggered when multiple actors converge on the same supply center, and the eventual payoff for any individual is intrinsically tied to their expectations of others' behaviors.
\end{itemize}

\subsection{StarCraft II (SC2) Macro-Decision Mining}
\label{app:sc2_details}

The SC2 pipeline focuses on high-level strategic pivots within continuous gameplay trajectories, mapping them to game-theoretic abstractions.

\begin{itemize}
    \item \textbf{Candidate Proposal:} The system analyzes trajectories for "heuristic peaks" or turning points where macro-action distributions shift significantly.
    \item \textbf{Window Validation:} Proposed windows are checked for feature consistency and outcome clarity.
    \item \textbf{Template Assignment via Intent Tags:} The system disambiguates the strategic nature of a window using intent signals:
    \begin{itemize}
        \item \textbf{Escalation Game:} Assigned if the sequence signals a rapid ramp-up in military commitment or production.
        \item \textbf{Sealed-Bid Auction:} Assigned if the data shows an "all-pay" competition for a resource (e.g., a contested expansion), where the investment is lost regardless of the outcome.
    \end{itemize}
\end{itemize}

\subsection{Occupation-Grounded Synthesis}
\label{app:synthesis_details}

This pipeline automates the creation of high-fidelity synthetic scenarios by grounding game-theoretic templates in professional contexts.

\subsubsection{Job Pool Construction}
To ensure diverse and realistic scenarios, a deduplicated "Job Pool" is generated from a large set of personas:
\begin{enumerate}
    \item \textbf{Extraction:} Job titles are parsed from raw persona descriptions.
    \item \textbf{LSH Deduplication:} A \textit{Locality Sensitive Hashing} (SimHash/MinHash) algorithm removes near-duplicate titles to ensure variety.
    \item \textbf{Embedding and Clustering:} The system uses an embedding model to cluster related roles, merging them into a final, distinct job pool $\mathcal{J}$.
\end{enumerate}

\subsubsection{Constrained Scenario Synthesis}
The dataset $\mathcal{D}_{syn}$ is populated through a controlled generation loop:
\begin{itemize}
    \item \textbf{Sampling:} For each iteration, the system samples a template $T$, a job $j$ from the pool, and environmental "controls" such as organizational stakes and time pressure.
    \item \textbf{LLM Generation:} An LLM generates a natural language scenario based on the sampled parameters.
    \item \textbf{Validation:} The output must pass a \textbf{Mechanical Check} (ensuring the game-theoretic logic of $T$ is present) and a \textbf{Grounding Check} (ensuring the scenario is authentic to the job $j$).
    \item \textbf{Final Deduplication:} A scenario-level deduplication pass is performed to eliminate semantic redundancy.
\end{itemize}

\subsection{Human annotation}
\begin{figure*}[htbp]
    \centering
    \includegraphics[width=0.8\textwidth]{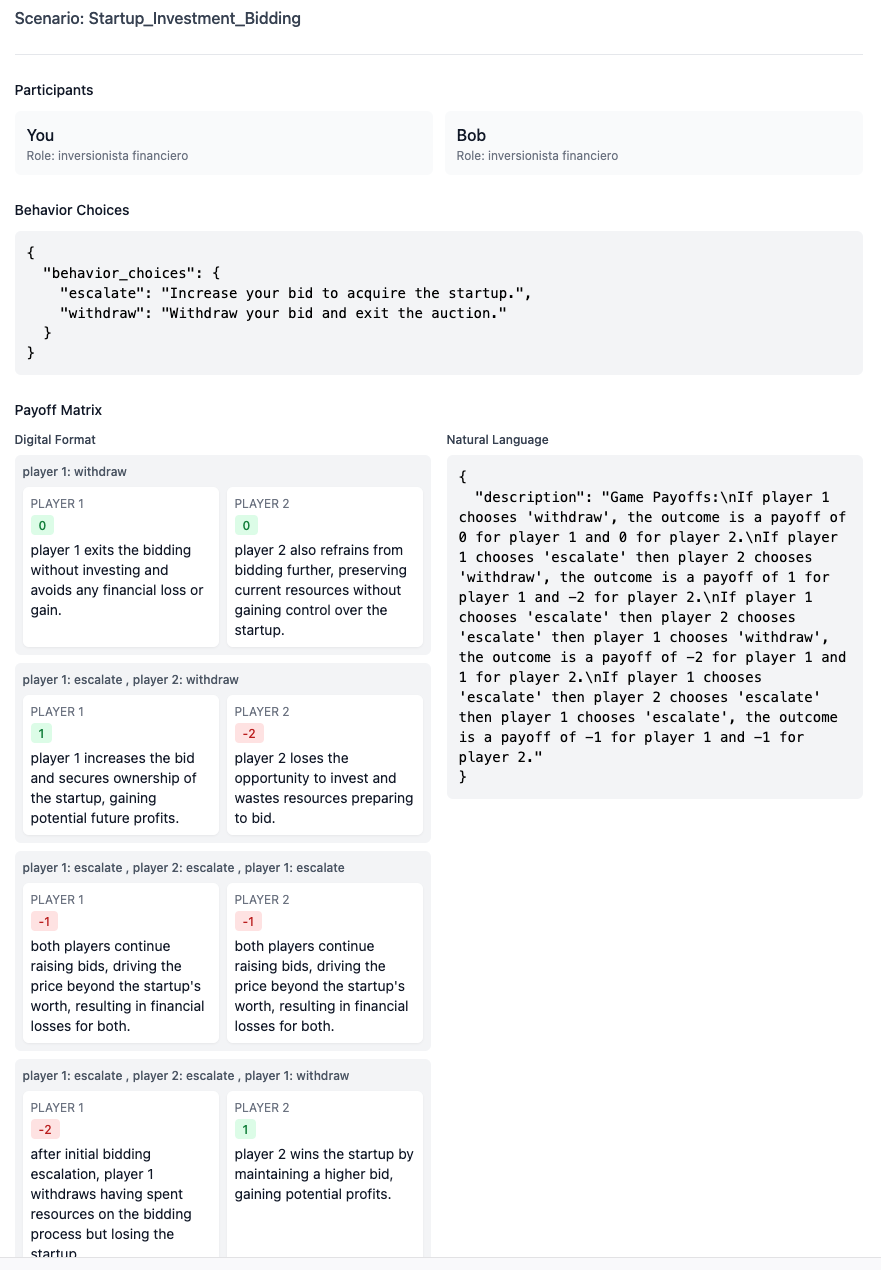}
    \caption{Page for human annotators} 
    \label{img:annotation_page}
\end{figure*}

With the extracted and synthesis dataset, two authors then annotate the samples independently. Annotators are prompted to check if a scenario follows the definition of a specific game theory template. The annotation interface is shown in  Figure~\ref{img:annotation_page}

%% file: appendix/stimulus_data_appendix.tex
\section{Emotion Eliciting Dataset}
\label{app:Emotion Eliciting Dataset}

\subsection{Emotion Eliciting Text Collection}
The textual component of our dataset was derived from the Crowd-enVent validation set, a collection of scenarios validated by crowd workers for emotional content. To ensure the reliability of these stimuli, we filtered the raw data to include only those entries with a validator agreement score of $\ge 0.8$ ($80\%$). The selection focused on six core emotions: Anger, Joy, Fear, Disgust, Sadness, and Surprise. Furthermore, we applied a two-step deduplication process, removing exact matches and filtering fuzzy duplicates with a textual similarity threshold of $0.9$. Finally, to standardize the presentation of these scenarios, we processed the high-agreement texts using GPT-5.2, transforming them into a clean, syntactically uniform version.

\subsection{Emotion-eliciting Image Generation}
The image generation pipeline was designed to produce high-quality visual stimuli that accurately reflect the emotional nuance of the source text.

\paragraph{Prompt Engineering}
To standardize the visual perspective, we employed a fixed prompt template: "draw a first perspective witness for the following case and show your \{emotion\}: \{text\}". This framing forces the model to adopt a subjective viewpoint, creating emotionally charged imagery rather than objective depictions.

\paragraph{Generation and Validation}
For each refined source text, OpenAI's DALL-E 3 model generated a set of candidate images, typically $k=3$ or $k=4$ variants. To select the optimal representation, we implemented an automated "Visual Critic" agent powered by GPT-5.2. This agent was presented with the original text, the target emotion, and the candidate pool, and was prompted to select the single image that most accurately and artistically represented the witness account. The system parsed the agent's decision to identify and preserve the winning image as the canonical visual stimulus.

\subsection{Emotion-eliciting Image Generation}
We developed a pipeline to generate visual stimuli corresponding to these refined texts. Using OpenAI's DALL-E 3, we created visual representations based on a standardized "first-person witness" prompt template designed to maximize emotional immersion. To ensure visual quality and semantic alignment, we generated multiple candidate images ($k=3$ or $k=4$) for each text and utilized a GPT-4o (Vision) agent to act as a "Visual Critic". This agent evaluated the candidates and selected the single best image based on composition and relevance to the original emotional scenario.

\subsection{Lexical and Syntactic Confounder Audit for Emotion Stimulus Dataset}
\label{app:Lexical and Syntactic Confounder}

To understand whether the emotion stimulus dataset can be distinguished easily by lexical and syntactic features, we used random forest for 5-fold cross-validation with a fixed set of eight shallow surface features:
\begin{itemize}
    \item \texttt{noun\_ratio}
    \item \texttt{verb\_ratio}
    \item \texttt{adjective\_ratio}
    \item \texttt{adverb\_ratio}
    \item \texttt{pronoun\_ratio}
    \item \texttt{determiner\_ratio}
    \item \texttt{adposition\_ratio}
    \item \texttt{coordinating\_conjunction\_ratio}
\end{itemize}

These features were chosen to emphasize shallow stylistic and surface-form properties rather than overt lexical framing or raw length cues.

Table~\ref{tab:surface_confound_accuracy} reports nested cross-validation results for the fixed eight-feature set.

\begin{table}[h]
\centering
\caption{Cross-validation accuracy for six-way emotion prediction using the fixed eight-feature shallow surface set.}
\label{tab:surface_confound_accuracy}
\begin{tabular}{lcccccc}
\hline
Dataset & $N$ & Features &  folds  & Acc by Chance & Acc by features \\
\hline
\texttt{Emotion Stimulus} & 241 & 8 & 5 & 0.1667 & 0.273639 $\pm$ 0.0470 \\
\hline
\end{tabular}
\end{table}

%% file: appendix/evaluate_metrics_appendix.tex
\section{The development of Behavior shift alignment metric}
\label{app:Behavior shift alignment}
\subsection{Systematic Review on Emotion decision impact }

We conducted a PRISMA 2020--informed systematic mapping review to synthesize evidence on how discrete emotions shift behavioral choices across canonical strategic ``game templates'' (Prisoner's Dilemma, Stag Hunt, escalation/commitment games, Trust Game, Ultimatum Game, sealed-bid auctions, and beauty-contest games) \citep{page2021prisma}. The review question was: \emph{How do experimentally induced or measured incidental emotions (happiness/positive mood, anger, disgust, fear, sadness, and surprise) systematically shift elicited decision behavior in commonly used economic games?}

\begin{table*}[ht]
\centering
\small
\begin{tabular}{lcccccc}
\toprule
\textbf{Game (Focal Behavior)} & \textbf{Happy} & \textbf{Anger} & \textbf{Disgust} & \textbf{Fear} & \textbf{Sad} & \textbf{Surp.} \\
\midrule
Prisoner's Dilemma (Cooperate) & $\uparrow$ & $\downarrow$ & $\downarrow$ & $\circ$ & $\circ$ & $\circ$ \\
Stag Hunt (Choose Stag)        & $\uparrow$   & $\circ$      & $\circ$      & $\downarrow$ & $\circ$ & $\circ$ \\
Escalation Game (Persist)      & $\circ$      & $\uparrow$   & $\circ$      & $\downarrow$ & $\circ$ & $\circ$ \\
Trust Game (Send/Trust)        & $\uparrow$   & $\downarrow$ & $\downarrow$ & $\downarrow$ & $\circ$ & $\circ$ \\
Ultimatum Game (Reject)        & $\downarrow$ & $\uparrow$   & $\uparrow$   & $\uparrow$   & $\uparrow$ & $\circ$ \\
Sealed-Bid (Overbidding)       & $\uparrow$   & $\uparrow$   & $\downarrow$ & $\downarrow$ & $\uparrow$ & $\uparrow$ \\
Beauty Contest (Depth)         & $\circ$      & $\downarrow$ & $\circ$      & $\circ$      & $\circ$ & $\circ$ \\
\bottomrule
\end{tabular}
\caption{Directional influence of emotions on game behavior: $\uparrow$ (increases behavior), $\downarrow$ (decreases behavior), $\circ$ (neutral/weak effect).}
\end{table*}

\paragraph{Eligibility criteria.}
We included empirical studies with (i) human participants, (ii) an experimental manipulation or measurement of a transient affective state (incidental emotion or mood, or experimentally elicited emotional expressions), and (iii) an incentivized or consequential decision task that either instantiated one of the target games or was isomorphic to a target game in payoff structure or strategic affordances (e.g., dyadic cooperation--defection, trust--reciprocity, costly punishment/rejection of unfairness, competitive bidding, or iterative strategic reasoning). We excluded purely correlational work without behavioral outcomes, clinical-only samples without choice tasks, and papers lacking sufficient methodological detail to identify the emotion induction/measurement or the decision template.

\paragraph{Information sources and search strategy.}
Searches were executed on January 6, 2026 using open scholarly indexing (e.g., Google Scholar) and publisher/author repositories. We combined game keywords with emotion keywords using Boolean strings. Examples: \texttt{("ultimatum game" OR bargaining) AND (anger OR disgust OR sadness OR fear OR happiness OR surprise)}, \texttt{("trust game") AND (emotion OR mood)}, \texttt{("beauty contest" OR guessing game) AND anger}, \texttt{("sealed-bid auction" OR first-price auction) AND emotion}, and \texttt{("stag hunt") AND affect*}. We complemented database-style keyword search with backward/forward citation chasing from highly-cited anchor papers in the appraisal-tendency tradition \citep{LernerKeltner2000,LernerKeltner2001} and from emotion-in-games papers identified during screening.

\paragraph{Study selection.}
Records were screened in two stages: title/abstract screening followed by full-text eligibility assessment. Because reporting standards and task labels vary across fields, we applied a ``template mapping'' rule: studies were retained if their decision problem could be mapped onto one of the target games by payoff/strategy structure even when authors used different labels (e.g., escalation of commitment tasks as escalation games; endowment/pricing tasks as auction-relevant valuation primitives).

\paragraph{Data extraction and coding.}
For each included study we extracted: emotion(s) manipulated/measured; induction/measurement method; task/game template; stakes/incentives; one-shot vs repeated interaction; primary behavioral dependent variables (e.g., cooperation rate, trust sent, reciprocity returned, rejection of unfair offers, bid level, strategic depth/guesses); and the direction of the reported emotion effect (increase/decrease/null/mixed). We coded effects as ``direct evidence'' when the target game was implemented, and as ``mapped evidence'' when a decision tendency supported an isomorphic implication for a target game (e.g., anger $\rightarrow$ greater punishment/rejection implies harsher responder behavior in Ultimatum settings).

\paragraph{Synthesis approach.}
Given heterogeneity in inductions, repeated vs one-shot designs, and outcome metrics, we used a narrative synthesis and evidence map rather than meta-analysis. When evidence was sparse or mixed for a specific Emotion $\times$ Game cell, we used appraisal-tendency theory \citep{LernerKeltner2000,LernerKeltner2001} to state a theoretically grounded directional prediction and explicitly labeled such cells as ``mapped'' or ``ambiguous'' rather than definitive.

\paragraph{Risk of bias and limitations.}
We qualitatively assessed internal validity using standard experimental design indicators (random assignment, manipulation checks, incentivization, preregistration where available, and attrition reporting). The principal limitations are: reliance on openly accessible indexing rather than subscription databases, potential publication bias, and unavoidable ambiguity when transporting emotion effects across superficially similar but strategically distinct implementations (e.g., one-shot vs repeated Prisoner's Dilemma).

%% file: appendix/irt-process.tex
\section{Data and Unit of Benchmark Analysis}
\label{latxIRT_analy}

This appendix briefly summarizes the IRT analysis used to screen the candidate pool and confirm the final released benchmark. It covers four points only: the analysis unit and response coding, the calibration procedure, the key validation criteria, and the main results from Stage~1 and Stage~2.

Each respondent profile corresponds to one model under one repeated neutral-condition run. Items were coded as either binary or ordered responses and were calibrated separately within each task family rather than on a single cross-game latent scale.

\section{Key Steps}

The analysis used a two-stage workflow.

\begin{enumerate}
  \item \textbf{Stage~1: candidate-pool calibration.} Binary task families were fit with Bayesian 2PL models, and ordered-response task families were fit with graded ordered logistic models, using neutral-condition responses only.
  \item \textbf{Stage~1 checks.} We evaluated posterior predictive fit, positive discrimination, ordered thresholds for ordinal items, and multi-seed stability across repeated refits.
  \item \textbf{Stage~2: final-benchmark confirmation.} After selecting the retained benchmark, we re-fit the same model families on retained items only and repeated the same checks.
  \item \textbf{Content audit.} Final retention also considered conceptual coverage and redundancy so that the benchmark would remain balanced and nonduplicative, not only psychometrically acceptable.
\end{enumerate}

\section{Validation Criteria}

The calibration was treated as acceptable when task families showed:
\begin{itemize}
  \item low posterior predictive error,
  \item positive item discrimination,
  \item correctly ordered thresholds for ordinal items, and
  \item stable parameter estimates across multiple random seeds.
\end{itemize}

The benchmark was calibrated within task family because the goal was not to estimate a single cross-game ability scale, but to capture task-specific behavioral tendencies under a common neutral condition.

\section{Stage~1 Results: Candidate Pool}

Stage~1 provided clear evidence that the candidate pool was psychometrically usable at the task level. All seven task families passed the basic validity checks: posterior predictive error was comfortably below the prespecified threshold, minimum discrimination remained positive in every family, and ordinal thresholds were well ordered where relevant.

The main limitation was seed stability. Although average parameter movement across seeds was moderate, the worst-case item movement exceeded the strict stability threshold in every task family. The practical interpretation is that Stage~1 was strong enough for item screening, curation, and broad task-level comparison, but not strong enough to justify heavy reliance on exact item-level parameter values from the full candidate pool.

\begin{table}[ht]
\centering
\resizebox{\textwidth}{!}{
\begin{tabular}{lccc}
\toprule
Task family & Mean abs. error & Minimum discrimination & Stage~1 conclusion \\
\midrule
Escalation Game & 0.0218 & 0.2952 & valid, but seed-unstable \\
Prisoners' Dilemma & 0.0217 & 0.2773 & valid, but seed-unstable \\
Stag Hunt & 0.0189 & 0.2822 & valid, but seed-unstable \\
Trust Game (trustee) & 0.0670 & 0.3859 & valid, but seed-unstable \\
Trust Game (trustor) & 0.0624 & 0.4056 & valid, but seed-unstable \\
Ultimatum Game (proposer) & 0.0292 & 0.3287 & valid, but seed-unstable \\
Ultimatum Game (responder) & 0.0205 & 0.2711 & valid, but seed-unstable \\
\bottomrule
\end{tabular}
}
\caption{Stage~1 summary for the candidate pool. All task families passed the core validity checks, but none met the strict multi-seed stability requirement.}
\end{table}

\section{Stage~2 Results: Final Limited Benchmark}

Stage~2 is the more important result for the released benchmark. After removing unstable or redundant candidates and retaining a smaller item set, we re-estimated each task family and repeated the same validity and reliability checks. The retained benchmark showed acceptable fit, positive discrimination, ordered thresholds where needed, and substantially improved seed stability.

\begin{table}[ht]
\centering
\resizebox{\textwidth}{!}{
\begin{tabular}{lrrrrrr}
\toprule
Task family & Retained items & Mean discr. & MAE & Min discr. & Max $\Delta a$ & Max $\Delta b$ \\
\midrule
Escalation Game & 12 & 1.18 & 0.018 & 0.52 & 0.24 & 0.19 \\
Prisoners' Dilemma & 12 & 1.23 & 0.018 & 0.49 & 0.27 & 0.22 \\
Stag Hunt & 10 & 1.38 & 0.016 & 0.56 & 0.21 & 0.18 \\
Trust Game (trustee) & 12 & 1.02 & 0.054 & 0.51 & 0.24 & 0.20 \\
Trust Game (trustor) & 12 & 1.00 & 0.050 & 0.54 & 0.22 & 0.19 \\
Ultimatum Game (proposer) & 10 & 0.98 & 0.024 & 0.47 & 0.20 & 0.17 \\
Ultimatum Game (responder) & 10 & 1.17 & 0.018 & 0.46 & 0.25 & 0.21 \\
\bottomrule
\end{tabular}
}
\caption{Stage~2 psychometric confirmation on the final limited benchmark.}
\label{tab:stage2_confirmation}
\end{table}

These results support the main claim needed for the appendix: the final retained benchmark is psychometrically defensible, whereas the larger candidate pool was useful primarily as a screening stage.

\section{Content Coverage and Redundancy}

Psychometric filtering was combined with a content audit to avoid a benchmark that was narrow or repetitive. Final retention therefore balanced four considerations: task-family coverage, role and scenario diversity, seed stability, and conceptual nonredundancy. When multiple items occupied essentially the same niche, preference was given to the item with stronger fit, better stability, and clearer coverage value.

This matters because a benchmark can fit well statistically while still overrepresenting a small part of the behavioral space. The content audit was used to ensure that the final benchmark remained broad enough to be useful while avoiding near-duplicate items.

\section{Conclusion}

The key conclusion is straightforward. Stage~1 showed that the candidate pool contained meaningful psychometric signal, but strict item-level stability was not yet strong enough to treat all candidate estimates as equally reliable. Stage~2 showed that, after selection and pruning, the final limited benchmark achieved good fit, positive discrimination, ordered thresholds, and improved stability across all task families. Accordingly, the released benchmark can be described as psychometrically supported and intentionally curated for both measurement quality and content coverage.